\documentclass{article}

\usepackage{dsfont}

\usepackage{arxiv}
\usepackage[utf8]{inputenc} 
\usepackage[T1]{fontenc}    
\usepackage{url}            
\usepackage{booktabs}       
\usepackage{amsfonts}       
\usepackage{nicefrac}       
\usepackage{microtype}      
\usepackage{lipsum}
\usepackage{amsmath}
\usepackage{amsthm}
\usepackage{amssymb}   
\usepackage{bigints}
\usepackage{bm}

\usepackage{diagbox}
\usepackage{makecell}
\usepackage{multirow}
\usepackage{caption}
\usepackage{subcaption}
\usepackage{graphicx}
\usepackage[ruled,vlined,linesnumbered]{algorithm2e}
\include{pythonlisting}
\usepackage{xcolor}
\usepackage{mathtools}
\usepackage{enumitem}

\usepackage{bbm}
\usepackage{hyperref}
\hypersetup{
    colorlinks=true,
    citecolor=black,
    linkcolor=black,
    filecolor=black,
    urlcolor=black,
}

\theoremstyle{definition}

\numberwithin{equation}{section}

\theoremstyle{plain}

\title{Scalable Bayesian optimization with high-dimensional outputs using randomized prior networks}

\author{
  Mohamed Aziz Bhouri \\
  Department of Mechanical Engineering \\
  and Applied Mechanics\\
  University of Pennsylvania\\
  Philadelphia, PA 19104 \\
  \texttt{bhouri@seas.upenn.edu } \\
  \And
  Michael Joly \\
  Raytheon Technologies \\
  \texttt{	michael.joly@rtx.com} \\
  \And
  Robert Yu \\
  Graduate Group in Applied Mathematics \\
  and Computational Science\\
  University of Pennsylvania\\
  Philadelphia, PA 19104 \\
  \texttt{ryu02@sas.upenn.edu} \\
  \And
  Soumalya Sarkar \\
  Raytheon Technologies \\
  \texttt{soumalya.sarkar@rtx.com} \\
  \And
  Paris Perdikaris \\
  Department of Mechanical Engineering \\
  and Applied Mechanics\\
  University of Pennsylvania\\
  Philadelphia, PA 19104 \\
  \texttt{pgp@seas.upenn.edu} \\
}

\begin{document}
\maketitle

\begin{abstract}

Several fundamental problems in science and engineering consist of global optimization tasks involving unknown high-dimensional (black-box) functions that map a set of controllable variables to the outcomes of an expensive experiment. Bayesian Optimization (BO) techniques are known to be effective in tackling global optimization problems using a relatively small number objective function evaluations, but their performance suffers when dealing with high-dimensional outputs. To overcome the major challenge of dimensionality, here we propose a deep learning framework for BO and sequential decision making based on bootstrapped ensembles of neural architectures with randomized priors. Using appropriate architecture choices, we show that the proposed framework can  approximate functional relationships between design variables and quantities of interest, even in cases where the latter take values in high-dimensional vector spaces or even infinite-dimensional function spaces. In the context of BO, we augmented the proposed probabilistic surrogates with re-parameterized Monte Carlo approximations of multiple-point (parallel) acquisition functions, as well as methodological extensions for accommodating black-box constraints and multi-fidelity information sources. We test the proposed framework against state-of-the-art methods for BO and demonstrate superior performance across several challenging tasks with high-dimensional outputs, including a constrained multi-fidelity optimization task involving shape optimization of rotor blades in turbo-machinery.

\end{abstract}

\section*{Highlights}
\begin{itemize}
	\item Development of a bootstrapped Randomized Prior Network (RPN) approach for Bayesian Optimization (BO).
	\item Extension of the proposed RPN-BO framework to the most general case of constrained multi-fidelity optimization.
	\item Formulation of appropriate re-parametrizations for evaluating common acquisition functions via Monte Carlo approximation, including parallel multi-point selection criteria for constrained and multi-fidelity optimization.
	\item Test of the proposed RPN-BO approach against state-of-the-art methods and demonstration of its superior performance across several challenging BO tasks with high-dimensional outputs.
\end{itemize}


\section{Introduction}

Solving global optimization problems involving high-dimensional input-output spaces and unknown or untraceable (black-box) objective functions is a ubiquitous task that many scientific and engineering fields face. These optimization problems naturally arise in various settings such as optimal control \cite{Lewis2012}, shape optimization \cite{Haslinger2003}, image denoising \cite{Fan2019}, finance \cite{Cornuejols2006} and aerodynamics \cite{Hicks1978}. Recently, several engineering tasks have been formulated as optimization problems; including optimizing radio cell towers coverage quality by optimizing their location \cite{Dreifuerst2020}, optimizing an optical intereferometer alignment in order to obtain light reflection without interference \cite{Sorokin2020}, and optimizing the design of a transonic compressor rotor with precompression \cite{Joly2019}.

In addition to the lack of knowledge or computational capacity to physically model the internal mechanisms mapping the high-dimensional input-output variables, other major sources of uncertainties include the stochasticity of the involved processes and the quality of the available data in terms of noise and sparsity \cite{Osband2022}. These problem settings can quickly result in intractable complexity, often referred to as the \textit{curse of dimensionality} \cite{maddox2021bayesian}. Given all these constraints, black-box optimization problems with high-dimensional variables are notoriously hard to solve, especially when various data with different fidelity levels must be considered, and/or additional constraints must be imposed to identify feasible and high-performance solutions.

\subsection{Related Work}
Previous strategies that address optimization problems with high-dimensional variables rely either on building computationally fast emulators that bypass the cost of expensive experiments \cite{Audet2000,Queipo2005,Forrester2009}, or on Bayesian Optimization (BO) approaches that account for the various sources of uncertainty \cite{Ryan2016,Lam2018}. For black-box function optimization, these surrogate models typically consists of auto-encoders \cite{Griffiths2020} and Gaussian Process (GP)-based formulations relying on Monte Carlo (MC) acquisition functions \cite{Wilson2018,astudillo2019bayesian,Balandat2020,Cakmak2020}. GPs are extensively used as surrogates thanks to their flexibility and analytically tractable epistemic uncertainty estimates. Most GP-based BO approaches select the next point(s) to evaluate an objective function by optimizing an acquisition function that is informed by the predictive distribution of the GP model. However, since these acquisition functions are often written as intractable integrals, MC approximation is used to enable fast sampling and evaluation of these quantities \cite{Wilson2018,astudillo2019bayesian,Balandat2020}. 

GP models typically scale poorly with the number of output variables, which greatly limits their applicability \cite{Swersky2013,Chowdhury2021}. Various recent works have focused on proposing improved GP-based BO approaches for problems with high-dimensional outputs. Among these efforts, one can mention composite BO approaches considering the setting of a differentiable objective function defined on the outputs of a vector-valued black-box function \cite{astudillo2019bayesian,Balandat2020,Uhrenholt2019,Maddox2021b,Astudillo2022}, contextual BO methods modeling a single function that varies across different environments or contexts \cite{Char2019,Feng2020,Krause2011}, and multi-objective BO approaches exploring a Pareto frontier across several objectives \cite{Daulton2020,Emmerich2006,Emmerich2011,Khan2002,Knowles2006}. Recently, Maddox {\it et al.} put forth a novel technique for exact multitask GP sampling that exploits Kronecker structure in the covariance matrices and leverages Matheron’s identity in order to perform BO with high-dimensional outputs using exact GP inference \cite{maddox2021bayesian}. 

Bayesian model updating and system identification techniques are widely used in various engineering fields including mechanical, aerospace and civil engineering \cite{Green2015,Crespo2021Calibration}. In these methods, a pre-defined model of the studied system must be available and the designed numerical algorithm searches for the optimal parameter setting that fits the model based on the observed data. These approaches include techniques based on Markov Chain Monte Carlo sampling \cite{Yang2020,dandekar2020bayesian,Wang2020added}, variational inference \cite{Sun2020} well designed distributions including staircase random variables and sliced normals \cite{Crespo2018Random,Crespo2021Robust,Colbert2020,Crespo2019quantification}, GP modeling \cite{Wenk2019}, sparsity-promoting prior distributions \cite{Nayek2021,Hirsh2022}, and even combinations of the last two techniques \cite{Bhouri2022}. Hence, these methods can also be seen as BO approaches where the Bayesian parameter inference is conducted via optimization routines that allow a finer tuning throughout this process but do not require newly acquired points. The BO context of this work consist of unknown or untraceable (black-box) objective functions for problems where an interpretable model of the studied system is not available. Therefore, the optimization is not conducted to find the best fit of the pre-defined model that best explains the observed data, but rather to find the best input that optimizes the unknown black-box objective function. These two categories of methods can be complementary based on the availability or absence of a pre-defined, and potentially interpretable, model.

Despite the aforementioned efforts to improve GP-based approaches in the context of BO with high-dimensional outputs, deep ensemble-based BO approaches have not been thoroughly explored yet. Deep ensembles provide an attractive alternative, especially in cases where  Bayesian inference is intractable or computationally prohibitive \cite{Lakshmin2017,Fort2019}. Indeed, for several practical applications, the resulting surrogate models have non-convex loss landscapes, leading to multi-modal posterior distributions that are hard to approximate or faithfully sample from \cite{Izmailov2018,Fort2019}. In addition, deep ensemble methods provide more flexibility against GP-based models in problems with high-dimensional input and/or outputs, since ensemble methods can leverage neural network architectures and capitalize on their ability to approximate functions and operators between high-dimensional spaces. However, the effectiveness of ensemble methods in decision making tasks has been put to question in recent years \cite{osband2018, Ciosek2020, Osband2022}.

Randomized prior networks (RPNs) were developed as an improvement over the classical deep ensembles by taking advantage of exiting prior belief to improve the model predictions in regions where limited or no training data is available \cite{osband2018}. This prior belief consists of  randomly initialized networks, where each single model is taken as a weighted sum of (1) a trainable neural network and (2) a second network with the same architecture but with fixed parameters that encode the prior belief and hence help capture the model's epistemic uncertainty. The model's ensemble captures uncertainties originating from the stochasticity of the modeled system and the quality of the limited data that is available. The asymptotic convergence to the true posterior distribution and the consistency of the RPNs have been proven theoretically under the assumption of inferring Gaussian linear models \cite{osband2018}. Additional theoretical studies proved the conservative uncertainty obtained with RPNs and its ability to reliably detect out-distribution samples \cite{Ciosek2020}. RPNs are also efficiently parallelizable and can be trained with a relatively small computational cost. In \cite{Osband2022} the authros showed that RPNs can match the posterior approximation quality of Hamiltonian Monte Carlo methods at a fraction of the computational cost, while outperforming GP-based approaches, mean-field variational approximations and dropout. These observations suggest RPNs as a prime candidate for tackling sequential decision making tasks by leveraging accurate approximation of joint predictive posteriors in order to effectively balance the exploration vs exploitation trade-off \cite{Ciosek2020,Osband2022}. 

\subsection{Main Contributions}
Drawing motivation from the aforementioned challenges and prior works, here we put forth the following contributions:
\begin{itemize}
    \item We propose a bootstrapped RPN approach for BO, where RPNs are not used to emulate the objective function directly, but rather to approximate the high-dimensional or functional outputs that a target objective function depends on.
    \item We extend the proposed RPN-BO framework to the most general case of constrained multi-fidelity optimization.
    \item We formulate appropriate re-parametrizations for evaluating common acquisition functions via Monte Carlo approximation, including parallel multi-point selection criteria for constrained and multi-fidelity optimization.
    \item We test the proposed RPN-BO approach against state-of-the-art methods and demonstrate superior performance across several challenging BO tasks with high-dimensional outputs.
\end{itemize}

\subsection{Organization}
Section \ref{sec:methods} provides a detailed presentation of the proposed methodology, while performance is assessed in section \ref{sec:res}. Our approach is compared against the state-of-the-art High-Order Gaussian Process method \cite{maddox2021bayesian} and the TuRBO approach \cite{Eriksson2019} across several high-dimensional Bayesian optimization benchmarks including the environmental model function example, the Brusselator PDE control problem, an optical intereferometer alignment problem, and a constrained multi-fidelity optimization case involving complex flow physics for identifying optimal rotor blade shapes. Section \ref{sec:cl} provides a summary of our findings and contributions and discusses remaining open questions and associated directions for future investigation.
 
\section{Methodology}
\label{sec:methods}

The proposed Bayesian Optimization (BO) approach relies on bootstrapped Randomized Prior Networks (RPNs) to provide a viable alternative to GP surrogates for high-dimensional BO problems. In section \ref{ss:BRPN-NC-SF}, the proposed framework is detailed for the case of non-constrained single-fidelity BO, while in section \ref{ss:BRPN-C-MF}, the proposed framework is extended to the most general case of constrained multi-fidelity BO. Any other problem setup in terms of constraint or fidelity level falls in between the complexity of these two cases where application of the proposed RPN-BO framework is straightforward.

\subsection{Non-Constrained Single-fidelity Bayesian Optimization}
\label{ss:BRPN-NC-SF} 

In the context of high-dimensional problems, the bootstrapped RPNs are built to map the input $\bm{x}\in\mathcal{X}$ of dimension $d_x$ to the output $\bm{y}$ of dimension $d_y$. The output is generally taken as a vector defined by the physical problem considered. Therefore, unlike most existing BO approaches, the RPN surrogate model is not used to emulate the objective function directly, but rather a physical quantity $\bm{y}$ that depends on the input $\bm{x}$ via a mapping noted $g(\cdot)$, and on which depends the objective function $f(\cdot)$. Therefore, the optimization problem can be written an follows:
\begin{equation}
    \min_{x\in\mathcal{X}\subset \mathbb{R}^d}f(g(\bm{x})),
\end{equation}
\noindent where RPNs are used to approximate the functional $g(\cdot)$ and $f(\cdot)$ is assumed to be known.

\subsubsection{Training Phase}
\label{sec:train}

RPNs are built by considering a function approximation:
\begin{equation}
    \label{equ:RPN}
    \hat{g}_{\theta,\gamma}(\bm{x}):=\hat{g}_\theta(\bm{x})+\beta \cdot \hat{g}_\gamma(\bm{x}),
\end{equation}
where $\hat{g}_\theta(\bm{x})$ is a neural network (NN) or any other machine learning surrogate model with trainable parameters $\theta$, while $\hat{g}_\gamma(\bm{x})$ is a so-called prior with exactly the same architecture as $\hat{g}_\theta(\bm{x})$, albeit with fixed, non-trainable parameters $\gamma$. The parameter $\beta$ is a scaling factor controlling the prior network variance and will always be taken equal to $1$ as justified in \cite{Yang2022Scalable}. 

The available training dataset consist of duplets of $\bm{x}$ and $\bm{y}$: $\mathcal{D}=\{(\bm{x}_i,\bm{y}_i),i=1,\ldots,N_d\}$. 

Multiple replicas of the  networks can be constructed by independent and random sampling of $\theta\sim p(\theta)$ and $\gamma\sim p(\gamma)$, leading to a deep ensemble model \cite{osband2018}. The distribution $p(\cdot)$ is defined based on standard initialization approaches for machine learning surrogate models. For instance, for multilayer perceptron (MLP) network, the Xavier Glorot initialization \cite{glorot2010understanding} provides a definition for the distribution $p(\cdot)$ for the weights and biases of fully connected neural networks. Similarly, existing work provide suitable distributions to initialize  nonlinear operators including for deep operator network (DeepONet) \cite{Lu2021} that can be used to define $p(\cdot)$. In addition to the prior incorporation of the prior in (\ref{equ:RPN}), we also resort to data bootstrapping in order to mitigate the uncertainty collapse of the ensemble method when tested beyond the training data points. Data bootstrapping (i.e. sub-sampling and randomization of the data each network in the ensemble sees during training), enables the efficient statistical estimation of well-calibrated confidence intervals \cite{osband2018}. Hence, for an ensemble of size $N_s$, each RPN member $\hat{g}_{\theta_i,\gamma_i}(\cdot)$, $i=1,\ldots,N_s$, is not trained on the full available dataset $\mathcal{D}$ but rather on a fraction $e$ ($0<e<1$) of the dataset $\mathcal{D}$. Each fraction is constructed by random and independent sampling from the global dataset $\mathcal{D}$. The fraction $e$ is fixed to $0.8$ in this work as justified in \cite{Yang2022Scalable}.

As a summary, each member $\hat{g}_{\theta_i,\gamma_i}(\cdot)$, $i=1,\ldots,N_s$, of the RPN ensemble is constructed by random sampling of $\theta_i\sim p(\theta)$ and $\gamma_i\sim p(\gamma)$. The non-trainable parameters $\gamma_i$ are kept fixed. The trainable parameters $\theta_i$ are learned by considering a randomly sampled fraction $e$ of the dataset $\mathcal{D}$ and matching the surrogate model $\hat{g}_{\theta_i,\gamma_i}(\cdot)$ predictions to the actual output values. In practice, we rely on stochastic gradient descent iterations to minimize an $L2$ loss function.

All networks in the ensemble can be trained in parallel via empirical risk minimization by leveraging for instance the vmap and pmap primitives in JAX \cite{bradbury2018jax}. This allows leveraging multi-threading on a single GPU, as well as parallelism across multiple GPU cards. Hence, multiple neural networks can be trained in parallel with effectively no additional computational cost compared to training a single network if enough GPU memory is available to train all neural networks simultaneously, as shown in table \ref{tab:time_cost}. Unlike Gaussian Process (GP) surrogates, this framework has no direct limitations with respect to scalability to high-dimensional inputs or outputs, nor intractable computational complexity as the number of training data is increased thanks to the proven expressiveness of deep neural networks in the context of inference involving high-dimensional inputs and/or outputs.

\begin{table}
\centering
\begin{tabular}{|c|c|}
\hline
$\#$ of NNs &  Training time (sec) \\
\hline
1 NN & $14.66\pm 0.56$\\ 
\hline
32 NNs & $14.61\pm 0.54$\\ 
\hline
64 NNs & $14.33\pm 0.63$\\ 
\hline
128 NNs & $14.27\pm 0.51$\\ 
\hline
\end{tabular}
\caption{{\em Time cost for RPNs training averaged over 10 independent runs for Ackley 2D problem  on a single NVIDIA Tesla P100 GPU}.}
\label{tab:time_cost}
\end{table}

\subsubsection{Prediction Phase}
\label{sec:pred}

The final output of a trained RPN ensemble is a collection of functional samples that resemble a Bayesian predictive posterior distribution. Each function in the output ensemble of size $N_s$ can be continuously queried at any new test input $\bm{x}_t$, yielding predictions with quantified uncertainty, and therefore directly facilitating downstream tasks like Bayesian optimization or active learning. Once an RPN surrogate is trained, an ensemble-based prediction can be obtained for the test point $\bm{x}_t$ by considering the fitted RPN members: $\{\hat{g}_{\theta_i,\gamma_i}(\bm{x}_t), i=1,\ldots,N_s\}$. Corresponding statistical estimates for the model's predictive mean and variance can be obtained via Monte Carlo (MC) ensemble averaging and by evaluating the forward pass of each network in the ensemble, which is also trivially parallelized and computationally efficient. 

\subsubsection{Backbone Architectures}

In this work, we considered neural networks as surrogate models $\hat{g}_\theta(\cdot)$ and $\hat{g}_\gamma(\cdot)$ to build the ensemble RPNs. In particular, multilayer perceptron (MLP) and DeepONet (DON) \cite{Lu2021} were used as neural networks' architectures. However, the proposed framework is general and independent from the particular surrogate model that is chosen. It can accommodate any surrogate model that is believed to be suited for the problem considered, including for instance other neural network architectures such as Convolution Neural Networks (CNN) \cite{OShea2015} , Recurrent Neural Networks (RNN) \cite{Sherstinsky2020} or Long Short-Term Memory (LSTM) networks \cite{Hochreiter1997}, or even different machine learning models such as Decision Trees (DTs) \cite{Quinlan1986} resulting in random forests \cite{Breiman2001}.

A multilayer perceptron (MLP) is built by considering a fully connected neural network of $N_l$ hidden layers (as shown in top left of figure \ref{fig:MF_RPN}). The corresponding forward pass for a duplet of input $\bm{x}$ and output $\bm{y}$ can be summarized as follows:
\begin{equation}\label{equ:frwd_MLP}
\begin{aligned}
    \bm{h_0} &=\bm{x} \ , \\ 
    \bm{h_i} &= F_{i-1}(\bm{W_{i-1}}\bm{h_{i-1}}+\bm{b_{i-1}}) \ , 1\leq i\leq N_l , \\
    \bm{h_{N_l+1}} &= \bm{W_{N_l}}\bm{h_{N_l}}+\bm{b_{N_l}} \ ,
\end{aligned}
\end{equation}
\noindent where $\bm{W_i}$ and $\bm{b_i}$ , $0\leq i\leq N_l$ are the trainable weights and biases respectively, and are tuned so that the output prediction $\bm{h_{N_l+1}}$ matches the data output $\bm{y}$ as closely as possible.

Thanks to the flexibility with respect to the surrogate model, the proposed RPN-based Bayesian optimization is also tested with an operator learning surrogate model consisting of the deep operator network (DeepONet) \cite{Lu2021}. DeepONet is designed to learn a functional mapping between a functional input $u$ and a functional output $G(u)$. The functional input is assumed to be evaluated at $d$ input sensors $m_1,\ldots,m_d$. The DeepONet architecture consist of a branch network mapping the evaluated functional input $u$ at the $d$ input sensors: $[u(m_1),\ldots,u(m_d)]$ to a $p$-dimensional output $b_1,\ldots,b_p$. It also contains a trunk network mapping the evaluation point $s$ considered for the functional output to a $p$-dimensional output $t_1,\ldots,t_p$ (Figure 1.d in \cite{Lu2021}). The branch and trunk outputs are then merged together to form the forward pass as follows:
\begin{equation}\label{equ:frwd_DON}
\begin{aligned}
    G(u)(s)=\sum\limits_{k=1}^p b_k[u(m_1),\ldots,u(m_d)]t_k(s) \ .
\end{aligned}
\end{equation}

Comparing with the problem setup defined above for the MLP network, the input $\bm{x}$ corresponds to the evaluation of the functional input $u$ at the $d$ input sensors $m_1,\ldots,m_d$: $\bm{x}\sim[u(m_1),\ldots,u(m_d)]$. The output $\bm{y}$ corresponds to the evaluation of the functional output $G(u)$ at $d_y$ output sensors $s_1,\ldots,s_{d_y}$: $\bm{y}\sim[G(u)(s_1),\ldots,G(u)(s_{d_y})]$. Similarly to the original DeepONet work \cite{Lu2021}, the branch and trunk networks are considered as fully connected neural network in this work.

\subsection{Constrained Multi-fidelity Bayesian Optimization}\label{ss:BRPN-C-MF}

More often than not, realistic design optimization problems involve several constraints that candidate designs should satisfy. Mathematically speaking such problems can be expressed in the form:
\begin{equation}\label{equ:CBO}
    \min_{\bm{x}\in\mathcal{X}\subset \mathbb{R}^D} f(g(\bm{x})) \ s.t.\ c_1(h_1(\bm{x})) \geq 0, \dots, c_K(h_K(\bm{x})) \geq 0, 
\end{equation}
where typically both the objective and constraints mappings, $g(\cdot)$ and $h_1(\cdot), \dots, h_K(\cdot)$ are black-box functions whose analytical form is unknown and only sample realizations can be collected.

Since we want to extend the proposed RPN-BO to the most general problem setup, we consider multi-fidelity (MF) objective and constraints. A low-fidelity (LF) RPN ensemble $\hat{g}_L$ is first built by training it to fit the low-fidelity vectorial output $\bm{y_L}$ for the objective mapping. $K$ low-fidelity RPN ensembles $\hat{h}_{L,k}$, $k=1,\ldots,K$ are also built to fit the low-fidelity vectorial outputs $y_{L,k}$ for the constraint functions. 

Each of $\hat{g}_L$ and $\hat{h}_{L,k}$, $k=1,\ldots,K$, refers to a complete surrogate model defined in Eq (\ref{equ:RPN}) as a weighted sum of a trainable and a non-trainable surrogate model. Their training and prediction processes are also the same as described in sections \ref{sec:train} and \ref{sec:pred} respectively.  Note that in some applications, the vectorial outputs for the objective and constraint mappings can be the same and only the scalar-output objective and constraint functions, $f(\cdot)$ and $c_{k}(\cdot)$, $k=1,\ldots,K$, differ. In such a case, only a single ensemble RPN is needed to be built for each fidelity level. The same output of the ensemble RPN is then taken as an input for the objective and constraint functions to evaluate the latter.

\begin{figure*}
\vskip 0.2in
\begin{center}
\centerline{\includegraphics[width=\columnwidth]{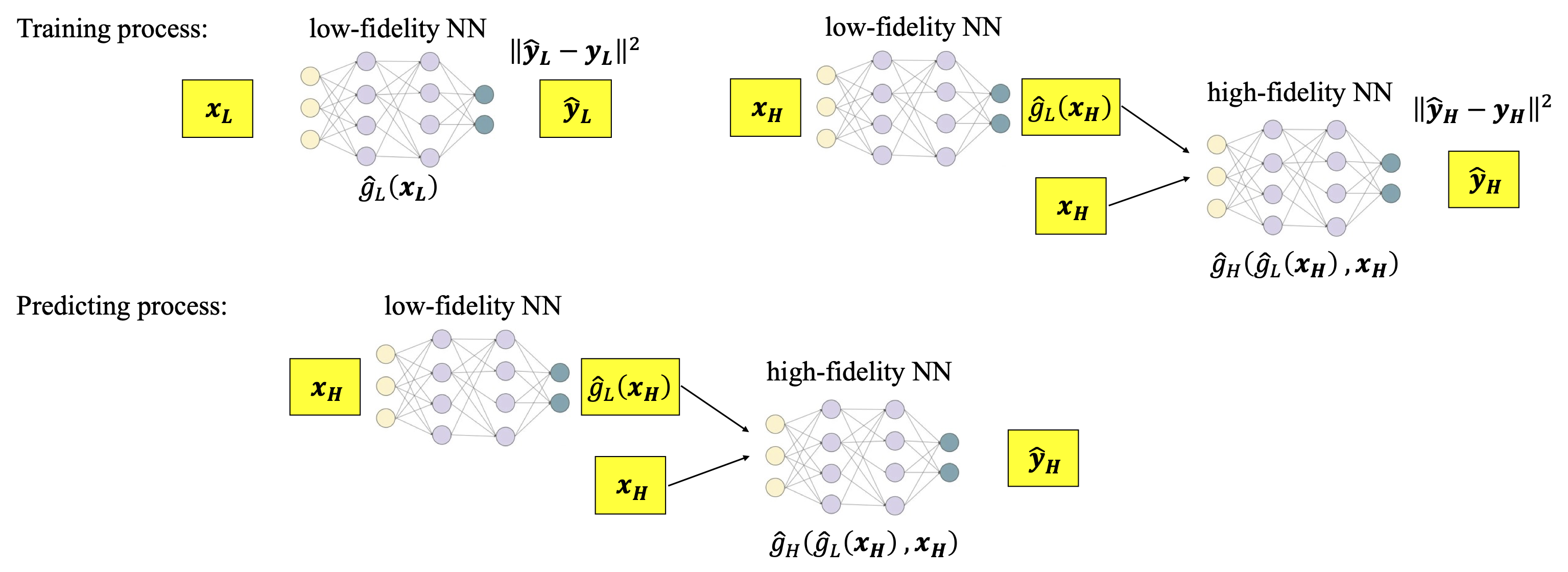}}
\caption{Multi-fidelity RPN formalism: NNs are used for illustrative pruposes. Other machine learning surrogate models can be considered.}
\label{fig:MF_RPN}
\end{center}
\vskip -0.2in
\end{figure*}

In order to consistently propagate uncertainty from low- to high-fidelity (HF) ensemble, each trained model $\hat{g}_L$ in the LF ensemble produces a prediction that is then used as input to train each member $\hat{g}_H$ of the HF ensemble, such that the output of the HF surrogate model for an input $\bm{x_H}$ is given by:
\begin{equation}
\label{equ:HF_RPN}
    \hat{y}_H=\hat{g}_H(\hat{g}_L(\bm{x_H}),\bm{x_H}),
\end{equation}
\noindent as summarized in figure \ref{fig:MF_RPN}. In addition to the consistent propagation of uncertainty from low- to high-fidelity model, the proposed architecture allows the vectorial outputs $\bm{y_L}$ and $\bm{y_H}$ of different fidelity levels to be of different dimensions. Such property is crucial for applications where different fidelity outputs are evaluated from models or experiments with different spatial and/or temporal discretizations. The constrained multi-fidelity optimization task involving shape optimization of rotor blades in turbo-machinery considered in this work and detailed in section \ref{ssec:shape_opt} is an example of these applications.

If constraints and objective functions depend on different vectorial outputs, the same training and prediction procedures are conducted for the multi-fidelity RPN ensemble models $\hat{h}_{H,k}$, $k=1,\ldots,K$ approximating the constraints' vectorial output. This formalism allows constructing probabilistic MF surrogates for each quantity of interest (e.g. objective, constraints), to be subsequently used in a Bayesian Optimization setting as detailed in the next section.
\subsection{Re-parameterized Monte Carlo Approximation of Multiple-Point (Parallel) Acquisition Functions}

Unlike Gaussian Processes (GP) surrogates, for RPNs (as for any other ensemble method) we can no longer derive analytical expressions for the posterior mean function and covariance in order to evaluate acquisition functions for sequential (or parallel) data acquisition in the context of Bayesian Optimization (BO). For this reason, we must resort to MC approximations. Different approaches for budget-constrained BO were recently proposed based on non-myopic acquisition functions \cite{Jiang2020,Jiang2020_2}. Here, unlike single acquisition criteria, multiple points can be selected as query points at each step based on maximizing the expected gain per multi-points. The non-myopic methods rely on dynamic programming and Markov decision process including  quasi-MC, common random numbers, and control variates in order to significantly reduce the computational burden associated with the optimization task \cite{Lee2020eff,Wu2019}. Although one-step/myopic approaches are not optimal, they offer cheaper computational cost. Some strategies for budget-constraint tasks were proposed where the computational cost is accumulated until the budget is exhausted \cite{pwf17}. 

In this work, we opt to rely on the multi-point acquisition functions based on MC sampling \cite{Wilson2018} and combine them with the bootstrapped RPN framework described above in order to perform BO. In the context of GP surrogates, these MC sampling approaches rely on the fact that many common acquisition functions $a(\cdot)$ can be expressed as the expectation of some real-valued function of the model output(s) at the design points, such that:
\begin{equation}\label{equ:acqE}
    a(\bm{X})=\mathbb{E}[a(\bm{\xi})|\bm{\xi}\sim P(f(g(\bm{X}))|\mathcal{D})],
\end{equation}
\noindent where $\bm{X}=(\bm{x}_1,\ldots,\bm{x}_q)$ and $P(f(g(\bm{X}))|\mathcal{D})$ is the posterior distribution of the function $f\circ g$ at the $q$ input points contained in $\bm{X}$ given the data $\mathcal{D}$ observed so far. 

Evaluating the acquisition function (\ref{equ:acqE}) therefore requires evaluating an integral over the posterior distribution. In most cases, this is analytically intractable. In particular, analytic expressions generally do not exist for batch acquisition functions that consider multiple design points jointly. An alternative is to use MC sampling to approximate these integrals. An MC approximation of $a$ at $\bm{x}$ using $N_s$ MC samples is:
\begin{equation}\label{equ:aMC}
    a(\bm{x})\approx \frac{1}{N_s}\sum\limits_{i=1}^{N_s} a(\bm{\xi}_i),
\end{equation}
\noindent where $\bm{\xi}_{i}\sim P(f(g(\bm{X}))|\mathcal{D})$. 

Reparameterization is needed in order to obtain an unbiased gradient estimator for the MC approximation of the acquisition function (\ref{equ:aMC}). This step produces samples that are interpreted differently making their differentiability with respect to a generative distribution’s parameters transparent \cite{Wilson2018}. Such re-parameterized MC approximation was proposed for different myopic maximal acquisition functions including: Probability of Improvement \cite{Viana2010,Kushner1964}, Simple Regret \cite{Azimi2010}, Expected Improvement (EI) \cite{Ginsbourger2010,Chevalier2013,Shahriari2016} and Lower Confidence Bound (LCB) \cite{Srinivas2010,Contal2013,Desautels2014} in the original work of \cite{Wilson2018}. Wilson et al. also developed re-parameterized MC approximations for two non-myopic acquisition functions: the Knowledge Gradient \cite{Frazier2008,Wu2016,Wu2017} and Entropy Search \cite{Hennig2012}. Re-parameterized MC approximations for other myopic acquisition functions such as Uncertainty Sampling \cite{Margatina2021} or Thompson Sampling (TS) \cite{Russo2018} are straightforward. However, such derivation can be slightly involving, yet still feasible, for other acquisition functions including Likelihood-Weighted LCB (LW-LCB) \cite{blanchard2021bayesian}, CLSF \cite{Costabal2019}, and Lower Confidence Bound with Constraints (LCBC) \cite{Noe2018,Zhang2022} acquisition functions. 

Using an ensemble method as surrogate model instead of Gaussian Processes allows a straightforward MC estimation of the acquisition functions thanks to the already available ensemble predictions $\{\hat{g}_{\theta_i,\gamma_i}(\cdot), i=1,\ldots,N_s\}$ whose estimation is trivially parallelized and computationally efficient as described in section \ref{sec:pred}. For instance, the re-parameterized MC approximation of the EI acquisition function can be derived by writing it as follows:
\begin{equation}\label{equ:acqEMC}
\begin{aligned}
    a_{\textrm{EI}}(\bm{X})\approx \frac{1}{N_s}\sum\limits_{i=1}^{N_s}\max\limits_{j=1,\ldots,q} \{ \max (\bm{\xi}_{i,j}-f^*,0) \} ,
\end{aligned}
\end{equation}
\noindent where $\bm{\xi}_{i}\sim P(f(g(\bm{X}))|\mathcal{D})$ and $f^*$ is the best scalar objective function value observed so far. Using the re-parametrization trick \cite{Wilson2018} in the context of the GP-based framework, the classical expected improvement criterion can be expressed as:
\begin{equation}\label{equ:qEIMC}
\begin{aligned}
    a_{\textrm{EI}}(\bm{X})\approx \frac{1}{N_s} \sum\limits_{i=1}^{N_s}\max\limits_{j=1,\ldots,q} \{ \max (\bm{\mu}(\bm{X})_j+(\bm{L}(\bm{X})\epsilon_i)_j-f^*,0) \},
\end{aligned}
\end{equation}
where $\epsilon_{i}\sim \mathcal{N}(0|I)$ and $\bm{\mu}(\bm{X})\in\mathbb{R}^{q}$ is the vector containing the GP's mean estimates for the scalar objective function $f\circ g$ evaluated at the $q$ points $(\bm{x}_1,\ldots,\bm{x}_q)=\bm{X}$, and  $\bm{L}\bm{L}^T=\bm{\Sigma}\in\mathbb{R}^{q\times q}$, where $\bm{\Sigma}$ is the GP's covariance estimates for the scalar objective function evaluated at the same $q$ points $(\bm{x}_1,\ldots,\bm{x}_q)$. 

In the context of the proposed RPN-BO, and since $\{f(\hat{g}_{\theta_i,\gamma_i}(\bm{x}_j)),1\leq i\leq N_s\}$ already provides a sampling from the posterior distribution $P(f(g(\bm{X}))|\mathcal{D})$, an MC estimate of the acquisition function using the RPN ensemble's prediction can be readily obtained as follows:
\begin{equation}\label{equ:acqEI_RPN}
\begin{aligned}
    a_{\textrm{EI}}(\bm{X})\approx \frac{1}{N_s} \sum\limits_{i=1}^{N_s}\max\limits_{j=1,\ldots,q} \{ \max (f(\hat{g}_{\theta_i,\gamma_i}(\bm{x}_j))-f^*,0) \} ,
\end{aligned}
\end{equation}
where $\theta_i\sim p(\theta) \ , \gamma_i\sim p(\gamma)$.

For sake of conciseness and clarity, we only detail here the re-parameterized MC approximation of the EI acquisition function. The re-parameterized MC approximations for the acquisition functions LW-LCB, CLSF and LW-LCBC are summarized in table \ref{tab:mc_acq}. These re-parameterizations are novel contributions that have not been previously explored in the BO literature, broadening the range of possible acquisition functions that can be considered for the proposed RPN-BO. The corresponding detailed derivations can be found in Appendix \ref{app_sec:reparam}.

\begin{table}
\centering
\begin{tabular}{|c|c|}
\hline
Abbr. &  Re-parameterized MC approximation for inputs  $(\bm{x}_1,\ldots,\bm{x}_q)$ \\
\hline
LW-LCB & $\frac{1}{N_s}\sum\limits_{i=1}^{N_s}\min\limits_{j=1,\ldots,q} \{ \mu(\bm{x}_j)-\sqrt{\kappa\pi/2} \ w(\bm{x}_j) \ |f(\hat{g}_{\theta_i,\gamma_i}(\bm{x}_j))-\mu(\bm{x}_j)| \}$\\ 
\hline
CLSF & $\frac{1}{N_s}\sum\limits_{i=1}^{N_s}\max\limits_{j=1,\ldots,q} \Big\{ \frac{\sqrt{\pi/2} \ w(\bm{x}_j) \ |f(\hat{g}_{\theta_i,\gamma_i}(\bm{x}_j))-\mu(\bm{x}_j)|}{ |\mu(\bm{x}_j)|^{\frac{1}{\kappa}} + \epsilon }
    \Big\}$ \\
\hline
LW-LCBC & ${\scriptstyle\Big[\frac{1}{N_s}\sum\limits_{i=1}^{N_s}\min\limits_{j=1,\ldots,q} \{ \mu(\bm{x}_j)-\delta-\sqrt{\kappa\pi/2} \ w(\bm{x}_j) \ |f(\hat{g}_{\theta_i,\gamma_i}(\bm{x}_j))-\mu(\bm{x}_j)| \}\Big] \times}$ \\
& ${\scriptstyle\prod_{k=1}^K \frac{1}{N_s}\sum_{i=1}^{N_s} \max\limits_{j=1,\ldots,q} \mathds{1}\{c_k(\hat{h}_{k,\theta^c_{k,i},\gamma^c_{k,i}}(\bm{x}_j))\geq 0\}}$\\ 
\hline
\end{tabular}
\caption{{\em Re-parameterized MC approximations:} $N_s$ is the RPN ensemble size, $\mu(\bm{x}_j)$, $1\leq j\leq q$, is the bootstrapped RPNs mean estimates of the scalar objective function evaluated at point $\bm{x}_j$, $\theta_i\sim p(\theta)$ and $\gamma_i\sim p(\gamma)$. $\kappa$ is a user-defined confidence parameter that controls the exploration versus exploitation trade-off and $w(\cdot)$ is a weight to regularize the uncertainty from the predicted posterior distribution \cite{Wilson2018,blanchard2021bayesian, blanchard2020informative, blanchard2020output}. For CLSF, $\epsilon$ is a small positive user-defined number. For LW-LCBC, $\delta=3$, $\theta^c_{k,i}\sim p(\theta^c_k)$ and $\gamma^c_{k,i}\sim p(\gamma^c_k)$. For LW-LCBC, $\hat{h}_{k,\theta^c_{k,i},\gamma^c_{k,i}}$, $1\leq k\leq K$, $1\leq i\leq N_s$, refer to the RPN surrogate models to approximate the constraint black-box functions $h_{k}$, $1\leq k\leq K$, with vectorial output, and in practice the sigmoid function is considered as a smooth indicator function $\mathds{1}$.}
\label{tab:mc_acq}
\end{table}

Algorithm \ref{alg:rpn_bo} summarizes the main steps to perform the proposed bootstrapped Randomized Prior Networks - Bayesian Optimization (RPN-BO). The algorithm's formulation is kept generic to accommodate all possible cases in terms of fidelity levels and constraint(s) existence.

\begin{algorithm}
\SetAlgoLined
Inputs: $\mathcal{D}$, $N_s$, $a(\cdot)$, $q$, $N$, $\epsilon$, surrogate model architecture \\ 
Outputs: $\mathcal{D}_f$, $f^*$ \\
$i = 0$ \\
$f^*$ = best objective value(s) using $\mathcal{D}$ points \\
$\mathcal{D}_f=\mathcal{D}$ \\
\While{$f^*>\epsilon$ and $i<N$}{$i=i+1$\\
Fit RPN ensemble(s) of size $N_s$ using $\mathcal{D}_f$\\
Determine $q$ point(s) $\bm{x}_1,\ldots,\bm{x}_q$ to acquire using fitted RPN(s) and MC approximation of $a(\cdot)$\\
Evaluate true vectorial output(s) for objective (and constraint(s) if any) correpsonding to $\bm{x}_1,\ldots,\bm{x}_q$ \\
Augment $\mathcal{D}_f$ with the newly acquired input and output point(s) \\
$f^*$ = best objective value(s), for different fidelity levels if any and among feasible solutions in case of constraint(s) based on $\mathcal{D}_f$\\}
\caption{Generic RPN-BO for a minimization task}
\label{alg:rpn_bo}
\end{algorithm}

\section{Results}\label{sec:res}
To assess the performance of the proposed bootstrapped Randomized Prior Networks - Bayesian Optimization (RPN-BO) framework and demonstrate its effectiveness, a comparison across several high-dimensional Bayesian optimization benchmarks was performed against the state-of-the-art High-Order Gaussian Process (HOGP) method \cite{maddox2021bayesian} and the TuRBO method \cite{Eriksson2019}. For RPN-BO, multilayer perceptron (MLP) and DeepONet (DON) \cite{Lu2021} were used as neural networks' architectures. $d_x$ denotes the dimension of the input design/control variables, and $d_y$ denotes the vectorial output's dimension. More details about hyperparameters can be found in Appendix \ref{app_sec:res}. All code and data accompanying this manuscript are publicly available at \url{https://github.com/bhouri0412/rpn\_bo}. 

\subsection{Environmental Model Function}

The Environmental Model Function problem is a well-known test problem in the literature of Bayesian calibration and models a chemical accident that results in a pollutant spill at two locations of a long and narrow holding channel \cite{bliznyuk2008bayesian,astudillo2019bayesian}. Using a first-order approach to model the substance's concentration within the channel, and modeling the latter as an infinitely long one-dimensional system with diffusion as the only transport mechanism, the chemical's concentration can be represented as:
\begin{equation}\label{equ:chemical}
\begin{aligned}
    c(s,t;M,D,L,\tau) = \frac{M}{\sqrt{4\pi Dt}}\exp\Big(\frac{-s^2}{4Dt}\Big)+\frac{\mathds{1}\{t>\tau\}M}{\sqrt{4\pi D(t-\tau)}}\exp\Big(\frac{-(s-L)^2}{4D(t-\tau)}\Big) \ , 
\end{aligned}
\end{equation}
\noindent where $M$ is the mass of pollutant spilled at each location, $D$ is the diffusion rate in the channel, $L$ is the location of the second spill and $\tau$ is the time of the second spill relative to the first one. The spatio-temporal pollutant's concentration is observed on a $3\times 4$ grid, resulting in the following concentration data: $\big\{c(s,t;M_0,D_0,L_0,\tau_0):(s,t)\in\{S\times T\}\big\}$, where $S=\{0,1,2.5\}$, $T=\{15,30,45,60\}$, and $(M_0,D_0,L_0,\tau_0)=(10,0.07,1.505,30.1525)$ are the underlying true values of these parameters and fixed equal to the values used in \cite{astudillo2019bayesian,maddox2021bayesian}.

The goal is to optimize the set of four parameters $(M,D,L,\tau)$ to achieve the true observed value by minimizing the mean squared error of the output grid:
\begin{equation}\label{equ:chemical_opt}
\begin{aligned}
    \frac{1}{12}\sum_{(s,t)\in S\times T} \big(c(s,t;M_0,D_0,L_0,\tau_0)-c(s,t;M,D,L,\tau)\big)^2.
\end{aligned}
\end{equation}

The input dimension for this optimization problem is $d_x=4$. The search domain is taken as fixed in \cite{astudillo2019bayesian,maddox2021bayesian} such that $M\in[7,12]$, $D\in[0.02,0.12]$, $L\in[0.01,3]$ and $\tau\in[30.01,30.295]$. The initial training dataset has a total of only $5$ points. For the RPN-BO method, RPNs learn the vectorial output of dimension $d_y=12$ corresponding to the $12$ grid values of the concentration. Then, the objective function (\ref{equ:chemical_opt}) is evaluated based on these inferred concentration values. 

\begin{figure}
\vskip 0.2in
\begin{center}
\centerline{\includegraphics[width=0.7\columnwidth]{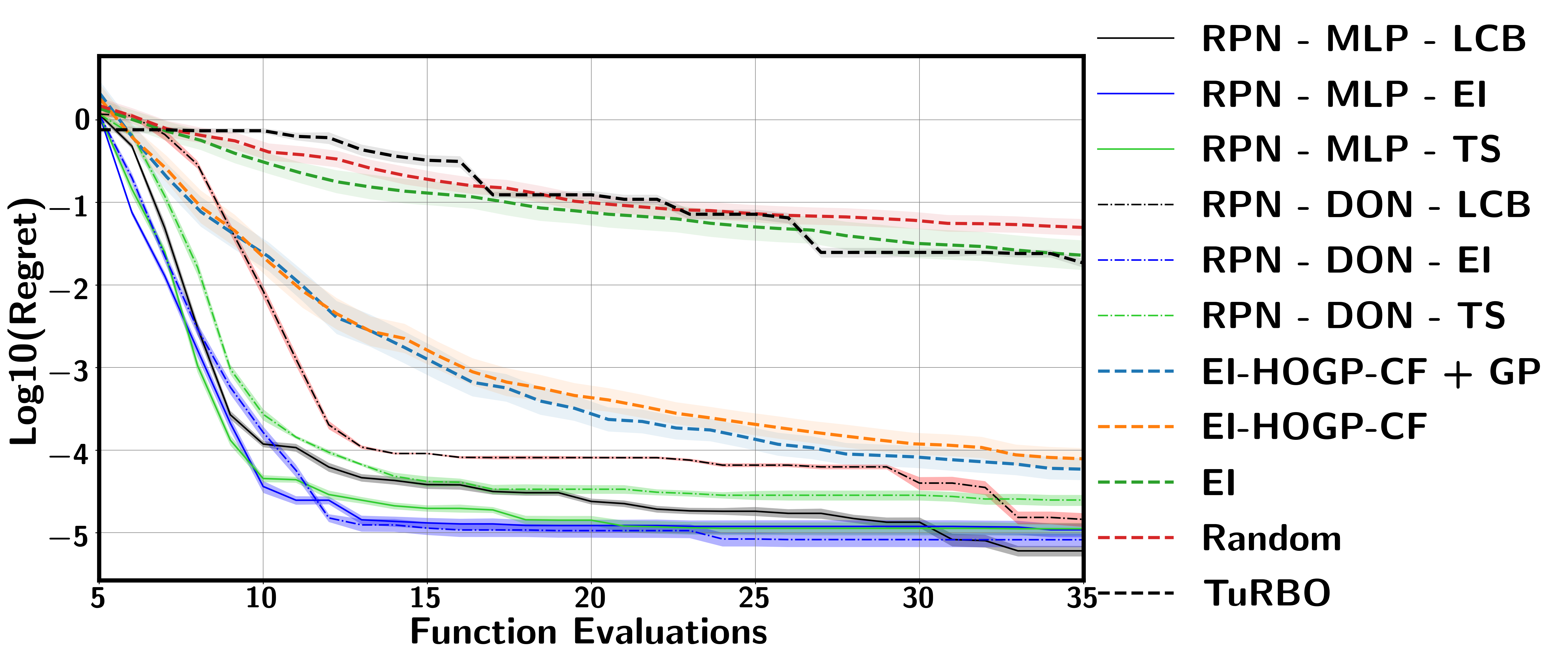}}
\caption{{\em High-dimensional Bayesian optimization for Environmental Model Function:} $d_x=4$, $d_y=12$.}
\label{fig:pollutants}
\end{center}
\vskip -0.2in
\end{figure}

Figure \ref{fig:pollutants} shows the optimization results averaged over $10$ independent random runs for different choices of acquisition functions and neural networks' architectures. The RPN-BO method using MLP with LCB, EI or TS acquisition functions has a significantly higher convergence rate compared to the HOGP method of \cite{maddox2021bayesian} and the TuRBO approach \cite{Eriksson2019}. The final objective value is also one order of magnitude lower for the MLP-based RPN-BO. The DON-based RPN-BO has a slightly lower initial convergence rate when using the LCB and TS acquisition functions compared to the MLP-based RPN-BO since DONs have more complex architectures compared to MLPs, and require further training points to be effectively fitted. Nonetheless, after acquiring only $3$ points for a training set of size $8$, all DON-based RPN-BO simulations show a significantly higher convergence rate compared to the HOGP and TuRBO methods, and reach a lower final objective value. 
\subsection{Brusselator PDE Control Problem}
\label{sec:PDE_control}
A larger experimental problem consists of optimizing two diffusivity and two rate parameters of a spatial Brusselator Partial Differential Equation (PDE) \cite{maddox2021bayesian}. To model a control problem of a dynamical system, the objective is set to minimize the weighted variance of the PDE solution. Similarly to the study performed in \cite{maddox2021bayesian}, we followed the example implementation given at \url{https://py-pde.readthedocs.io/en/latest/examples_gallery/pde_brusselator_expression.html#sphx-glr-examples-gallery-pde-brusselator-expression-py} in order to solve in \texttt{py-pde} \cite{zwicker2020py} the following Brusselator PDE with spatial coupling:
\begin{equation}
    \begin{aligned}\label{equ:brusselator}
        \partial_t u &= D_0\nabla^2u + a - (1+b)u + vu^2, \\
        \partial_t v &= D_1\nabla^2v + bu - vu^2 \ ,
    \end{aligned}
\end{equation}
\noindent where $D_0$ and $D_1$ are the diffusivity and $a$ and $b$ are the rate parameters to optimize. 

The weighted variance of the PDE (\ref{equ:brusselator})'s solution, as defined in \cite{maddox2021bayesian}, is taken as objective function to be minimized in order to guarantee a fair comparison between the different BO approaches. 
The Brusselator PDE (\ref{equ:brusselator}) is solved on a $64\times 64$ grid, producing output solutions of size $2\times 64\times 64$. The input dimension for this example is $d_x=4$. The search domain is taken as chosen in \cite{maddox2021bayesian} such that $a\in[0.1,5]$, $b\in[0.1,5]$, $D_0\in[0.01,5]$ and $D_1\in[0.01,5]$. The initial training dataset has a total of only $5$ points. For the RPN-BO methods, RPNs learn the vectorial output of dimension $d_y=8192$ corresponding to the $2$-dimensional solution to the spatial Brusselator PDE (\ref{equ:brusselator}) discretized over the $64\times 64$ grid. Then the objective function defined as the weighted variance of this solution is evaluated based on these inferred solution values.

\begin{figure}
\vskip 0.2in
\begin{center}
\centerline{\includegraphics[width=0.7\columnwidth]{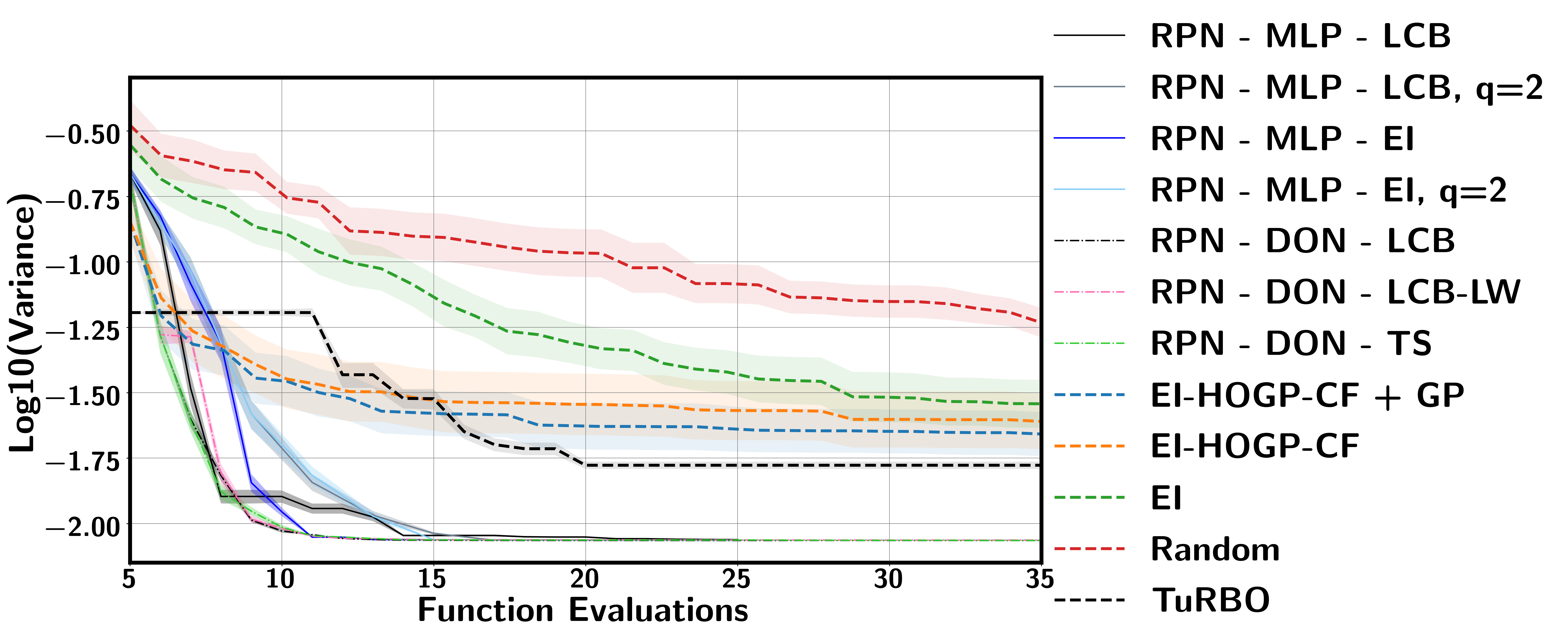}}
\caption{{\em High-dimensional Bayesian optimization for Brusselator PDE Control Problem:} $d_x=4$, $d_y=8192$.}
\label{fig:Brusselator}
\end{center}
\vskip -0.2in
\end{figure}

Figure \ref{fig:Brusselator} shows the optimization results averaged over $10$ independent random runs for different choices of acquisition functions and neural networks' architectures. Here, the LCB, LCB-LW, TS and EI acquisition functions were chosen for the RPN-BO approaches. These results demonstrate that RPN-BO is robust with respect to the acquisition function and neural networks' architecture. Indeed, for any choice of acquisition function and neural networks' architecture (MLP or DeepONet), the RPN-BO does not only converge faster to the optimal solution compared to the HOGP and TuRBO methods, but it also finds a significantly better optimum which is half an order of magnitude smaller than the minimum reached by HOGP and TuRBO approaches. The convergence results are similar for both neural networks' architectures. Unlike the environmental model function problem, the DON-based RPN-BO approach does not show slower convergence for the first iterations compared to the MLP-based RPN-BO, since DON architectures are designed to learn functionals. These results show the ability of the proposed RPN-BO to find better optimal solutions to the optimization problem with a faster convergence rate. This is due to the inherent ability of neural networks to better approximate high-dimensional spaces compared to GP-based approaches, including the HOGP methods, even with significantly sparse training datasets.

Figure \ref{fig:Brusselator} also shows the convergence results obtained with $q=2$ using the MLP-based RPN-BO with the LCB and EI acquisition functions. For both simulations, the RPN-BO converges slightly slower but reaches the same optimal value as when acquiring a single point at each optimization step. This behavior is expected since for higher values of $q$, the RPN-BO algorithm is asked to acquire more points while being trained on the same dataset size compared to the case of lower values of $q$. This tends to slow the overall optimization convergence. However, the total number of ensemble neural networks to train is lower for higher values of $q$. Hence, the optimization algorithm has a lower overall computational cost for higher values of acquired points per optimization step $q$. The latter should be chosen as a trade-off between the desired optimization convergence rate and acceptable optimization computational cost.
\subsection{Optical Interferometer Alignment}
The third problem consists of aligning an optical intereferometer by optimizing the coordinates of two mirrors so that the intereferometer reflects light without interference \cite{Sorokin2020,maddox2021bayesian}. The optical system consist of $16$ different interference patterns, each consisting of a $64\times 64$ image, resulting in an output tensor of shape $16\times64\times64$. The simulator implementation is available at \url{https://github.com/dmitrySorokin/interferobotProject}. The objective is to maximize the visibility $V$ that depends on the $16$ different interference patterns and defined as follows:
\begin{align}
    \label{eq:intensity}{\rm Intensity}_t &= \sum_{i,j=1}^{64} \exp\{-(i/64-0.5)^2-(j/64-0.5)^2\}*I_t \ , \\
    \label{eq:Imax}I_{\rm max} &= \underset{t=1,\ldots,16}{{\rm LogSumExp}} ({\rm Intensity}_t)  \ , \\
    \label{eq:Imin}I_{\rm min} &= -\underset{t=1,\ldots,16}{{\rm LogSumExp}}(-{\rm Intensity}_t)  \ , \\
    \label{eq:V} V &= (I_{\rm max}-I_{\rm min}) / (I_{\rm max}+I_{\rm min})
\end{align}
\noindent where $I_t$ is the $t$-th $64\times64$ image output of the model among the $16$ different interference patterns.  The input dimension for this example is $d_x=4$ consisting of the simulator values. The search domain is taken equal to $[-1,1]^4$ as chosen in \cite{maddox2021bayesian}. The initial dataset has a total of $15$ points. For the RPN-BO method, RPNs learn the vectorial output of dimension $d_y=65536$, corresponding to the $16$ $64\times 64$ images of the interference patterns. Then the objective function defined as the visibility $V$ (\ref{eq:V}) is evaluated based on these inferred images.

\begin{figure}
\vskip 0.2in
\begin{center}
\centerline{\includegraphics[width=0.7\columnwidth]{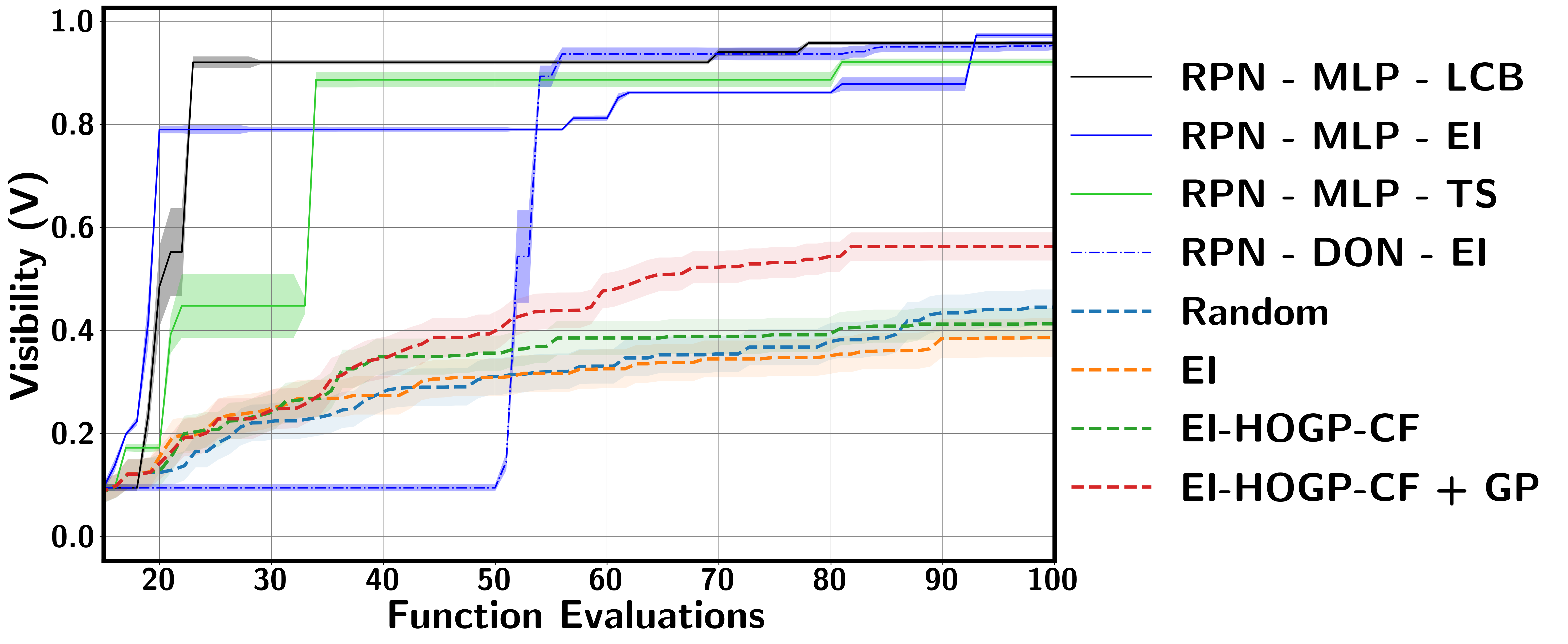}}
\caption{{\em High-dimensional Bayesian optimization for Optical Interferometer:} $d_x=4$, $d_y=65536$.}
\label{fig:optics}
\end{center}
\vskip -0.2in
\end{figure}

Figure \ref{fig:optics} shows the optimization results averaged over $5$ independent random runs obtained with the MLP- and DON-based RPN-BO method using the LCB, EI and TS acquisition functions. Similarly to the previous examples, The MLP-based RPN-BO displays the highest convergence rate. Although the DON-based RPN-BO shows a non-convergence for the first optimization iterations, once the training dataset is large enough ($\sim 50$ points), we observe a steep convergence similar to the convergence obtained with the MLP-based RPN-BO for the first acquired points. This can be explained by the more complex architecture of DeepONets compared to MLP, which requires more training points. Independently form the neural nets' architecture, the RPN-BO approach always showed a significantly faster convergence compared to the HOGP method which displayed a rather pronounced stagnation for the returned optimal solution. Indeed, the RPN-BO simulations always returned optimal objectives that have a visibility close to the physical limit of $1$, while the HOGP optimal solution does not exceed $0.6$. The RPN-BO method is also robust with respect to the acquisition function, since for each of the LCB, EI and TS criteria, the MLP-based RPN-BO was always capable of finding an optimum that is close to the physical limit of $1$.

\subsection{Constrained Multi-fidelity Optimization of Compressor's Blades Shape Design}\label{ssec:shape_opt}
\begin{figure}
\vskip 0.2in
\begin{center}
\centerline{\includegraphics[width=.5\columnwidth]{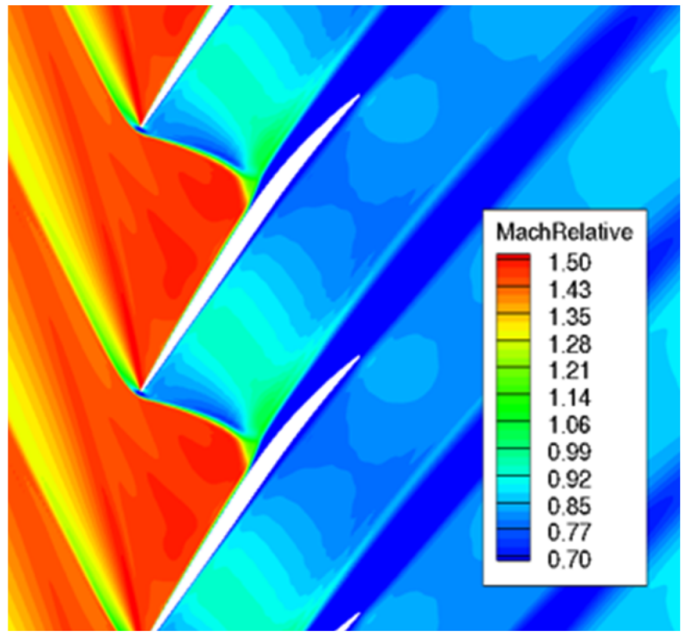}}
\caption{{\em Relative Mach number filed at $70\%$ span for baseline design.}}
\label{fig:blade}
\end{center}
\vskip -0.2in
\end{figure}

The last experiment consists of a constrained multi-fidelity optimization of a compressor's blades shape design for transonic airfoils as shown in the illustrative figure \ref{fig:blade} adapted from \cite{Joly2019}. The input dimension for this example is $d_x=8$ consisting of the compressor's blades shape parameters, comprising $3$ offset points along the thickness distribution and $5$ offset points along the camber distribution. The objective of each compressor stage is to increase the total pressure from inlet to outlet while minimizing the required mechanical work. Thus, the efficiency is a critical design objective for a compressor stage and can be written as:
\begin{equation}\label{equ:efficiency}
    \eta=\frac{\Big(p_2/p_1\Big)^{(\gamma-1)/\gamma}-1}{T_2/T_1-1},
\end{equation}
for an adiabatic regime and assuming an ideal gas. $p_1$, $p_2$, $T_1$ and $T_2$ are the mass-flow averaged stagnation pressures and temperatures at the inlet and outlet, respectively. $\gamma$ is the isentropic expansion factor.

The stagnation pressure ratio $r=p_2/p_1$ is a prime design target for engine sizing. The optimization problem of interest is formulated to improve the efficiency of a baseline design taken as $70\%$ span section of the NASA rotor $37$, while at least maintaining the corresponding baseline $r_0$. Hence the objective function is taken as $-\eta$ to minimize, while maintaining $p_2/p_1\geq r_0$ which defines the constraint function for BO. Data acquisition of the objective and constraint requires solving the fluid dynamics problem of transonic flow around the compressor blades. The objective and constraint functions are computed via a Computational Fluid Dynamics (CFD) solver, using Raytheon Technologies' in-house CFD solver UTCFD, which solves the 2D compressible Reynolds Averaged Navier-Stokes (RANS) equations using the $k-\omega$ turbulence model. 

\begin{figure}
\vskip 0.2in
\begin{center}
\centerline{\includegraphics[width=0.7\columnwidth]{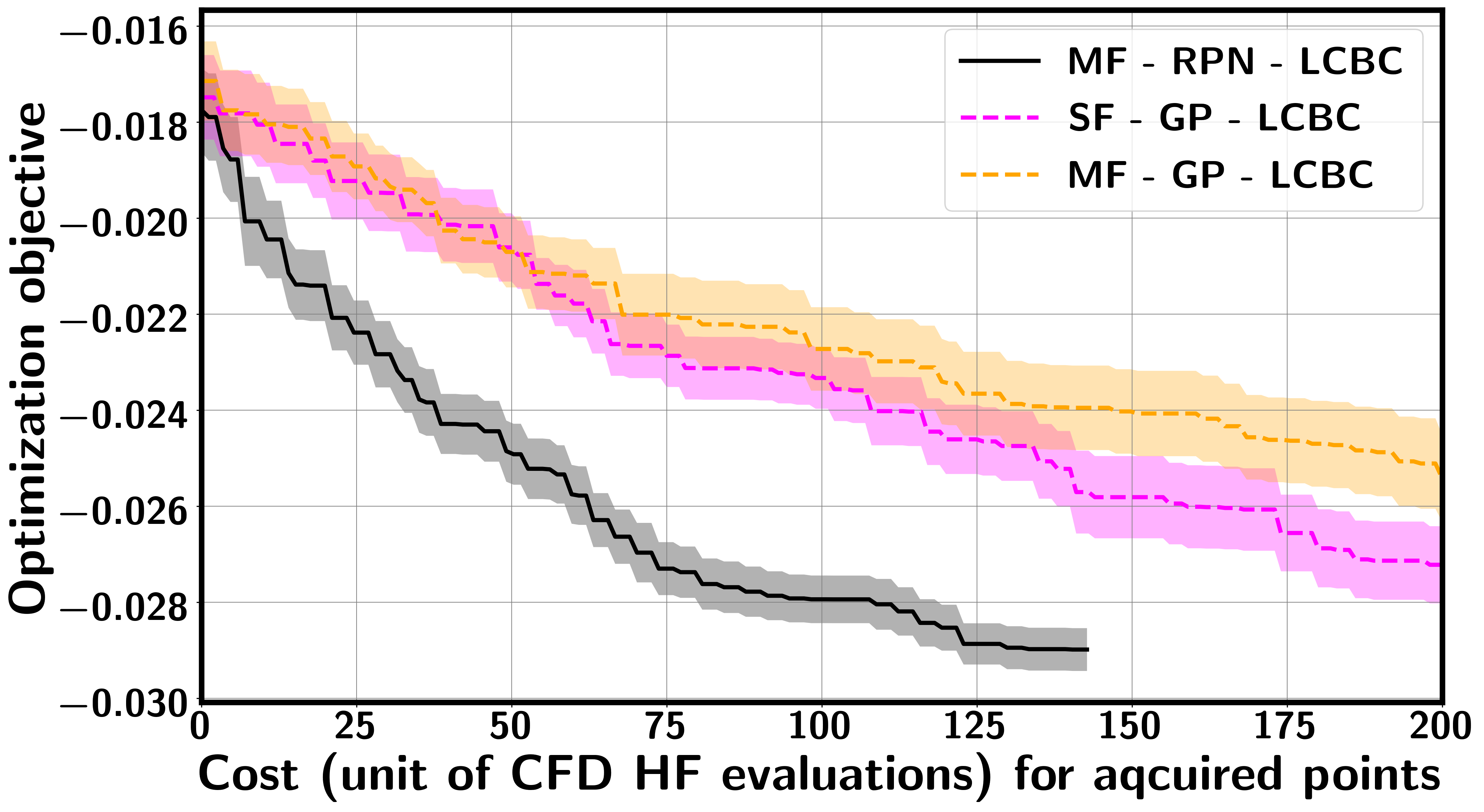}}
\caption{{\em Constrained Multi-fidelity Optimization for Compressor's Blades Shape Design:} convergence of normalized $-\eta$.}
\label{fig:RPN_compressor}
\end{center}
\vskip -0.2in
\end{figure}

Different levels of fidelity are generated using different levels of grid discretizations. The low- and high-fidelity models contain $10^4$ and $2.5\times10^4$ grid points, and take $\sim1.5$ and $\sim8$ minutes to run on $4$ cores, respectively.
Figure \ref{fig:RPN_compressor} shows the optimization results averaged over $20$ independent random runs and obtained with single-, multi-fidelity GP-BO and multi-fidelity MLP-based RPN-BO approaches using the LCBC acquisition function for the objective and the CLSF criterion for the constraint. Similarly to the previous examples, the MLP-based RPN-BO displays a higher convergence rate and is capable of finding a better optimal solution compared to the GP-based BO approaches. This behavior can be explained by the capacity of neural networks ensembles to emulate relatively complex and high-dimensional functionals. On the other hand, GP models typically scale poorly with the complexity of the inferred input/output dependencies. Besides, MF RPNs shows promises in leveraging LF data to accelerate optimization. Since the sampling ratio used in this numerical example is 1HF/1LF, further cost benefits are expected with multi-point sampling in a cost-aware optimization setting.

\subsection{Computation time}

All RPN-BO runs were conducted on a single NVIDIA Tesla P100 GPU. Acquisition functions optimization is always performed with 500 restarts in all numerical examples and the best optimal solution among the 500 ones that are found is taken as final solution based on the predicted acquisition function evaluation with the surrogate RPN.

Tables \ref{tab:time_cost_pollutant}, \ref{tab:time_cost_pde} and \ref{tab:time_cost_optics} show the time cost for the different RPN-BO settings considered that is needed determine a newly acquired point for the Environmental Model Function, the Brusselator PDE Control Problem, and the Optical Interferometer Alignment problem respectively, using the chosen hyperparameter setting detailed in Appendix \ref{app_sec:res}. For the Constrained Multi-fidelity Optimization of Compressor’s Blades Shape Design, the average time cost to train the RPN surrogate model takes $10.4 \ s$, while the RPN-based acquisition function optimization is performed within $11.9 \ s$. 

\begin{table}
\centering
\begin{tabular}{|c|c|}
\hline
RPN-BO setting &  Time cost \\
\hline
MLP-LCB & $19.9 \ (13.3)$\\ 
\hline
MLP-EI & $14.6 \ (8.0)$\\ 
\hline
MLP-TS & $18.4 \ (11.8)$\\ 
\hline
DON-LCB & $19.9 \ (12.4)$\\ 
\hline
DON-EI & $15.1 \ (7.6)$\\ 
\hline
DON-TS & $19.1 \ (11.6)$\\ 
\hline
\end{tabular}
\caption{{\em Average time cost in seconds for different RPN-BO settings to determine a newly acquired point for the Environmental Model Function on a single NVIDIA Tesla P100 GPU}. Acquisition function optimization is conducted with 500 restarts. The reported time cost includes both (1) RPN training and (2) RPN-based acquisition function optimization. The timing between parentheses indicates the time cost for the RPN-based acquisition function optimization.}
\label{tab:time_cost_pollutant}
\end{table}

\begin{table}
\centering
\begin{tabular}{|c|c|}
\hline
RPN-BO setting &  Time cost \\
\hline
MLP-LCB & $114 \ (32.7)$\\ 
\hline
MLP-LCB, $q=2$ & $64.2 \ (23.6)$\\ 
\hline
MLP-EI & $132.8 \ (51.5)$\\ 
\hline
MLP-EI, $q=2$ & $65.8 \ (25.1)$\\ 
\hline
DON-LCB & $208 \ (38.5)$\\ 
\hline
DON-LCB-LW & $219.5 \ (50)$\\ 
\hline
DON-TS & $208 \ (38.5)$\\ 
\hline
\end{tabular}
\caption{{\em Average time cost in seconds for different RPN-BO settings to determine a newly acquired point for the Brusselator PDE Control Problem on a single NVIDIA Tesla P100 GPU}. Acquisition function optimization is conducted with 500 restarts. The reported time cost includes both (1) RPN training and (2) RPN-based acquisition function optimization. The timing between parentheses indicates the time cost for the RPN-based acquisition function optimization.}
\label{tab:time_cost_pde}
\end{table}

\begin{table}
\centering
\begin{tabular}{|c|c|}
\hline
RPN-BO setting &  Time cost \\
\hline
MLP-LCB & $180.7 \ (81.3)$\\ 
\hline
MLP-EI & $162 \ (62.6)$\\ 
\hline
MLP-TS & $161.2 \ (61.8)$\\ 
\hline
DON-EI & $247.7 \ (77.6)$\\ 
\hline
\end{tabular}
\caption{{\em Average time cost in seconds for different RPN-BO settings to determine a newly acquired point for the Optical Interferometer Alignment problem on a single NVIDIA Tesla P100 GPU}. Acquisition function optimization is conducted with 500 restarts. The reported time cost includes both (1) RPN training and (2) RPN-based acquisition function optimization. The timing between parentheses indicates the time cost for the RPN-based acquisition function optimization.}
\label{tab:time_cost_optics}
\end{table}

Calling the physical model takes $0.67 \ s$, $12.8 \ s$ and $2.64 \ s$ for the Environmental Model Function, the Brusselator PDE Control Problem, and the Optical Interferometer Alignment problem respectively, while it takes $9.5$ minutes for the Compressor's Blades Shape Design problem. Hence, if the acquisition function is optimized only once (not with the 500 restarts as considered in this work for robustness), then calling the physical model is always more expensive than optimizing the acquisition function for all neural network architectures considered (MLP and DeepONet) and all acquisition functions that were tested across all problems. Even if we consider the total cost of all 500 restarts that were used in the acquisition function optimization, then calling the physical model becomes more expensive than optimizing the acquisition function as we consider more complex physical systems as is the case for the Compressor's Blades Shape Design problem.

For all examples considered, we observed a sufficiently fast convergence. For instance, 7 acquisition steps are indeed enough for the Environmental Model Function example using the MLP-based RPN-BO instead of the 30 steps considered for illustration purposes as seen in figure \ref{fig:pollutants}. Similarly, 5 acquisition steps are enough for Brusselator PDE Control Problem using the MLP-based RPN-BO instead of the 30 steps considered for illustration purposes as observed in figure \ref{fig:Brusselator}  and 10 acquisition steps instead of 85 are enough for the optical interferometer example as seen in figure \ref{fig:optics}. Hence the fast convergence of the RPN-BO does help remedy reducing the overall computational cost associated with the BO task of finding the optimal solution.

Relatively simple NN architecture and early stopping for training were also favored to avoid over-fitting in the case of sparse training data regime considered as detailed in Appendix \ref{app_sec:res}. These hyper-parameters choices also favor a lower computational cost. In addition, the acquisition function optimization is also needed at each iteration for GP-based BO approaches. In addition, the latter require computing, storing and decomposing the covariance matrix which scales poorly with the size of the output. Hence in the context of high-dimensional outputs, the better capability of approximating them at a lower cost using RPNs is a crucial factor in lowering the overall computational cost when using RPN-BO compared to GP-based BO approaches for the same optimization performance.

Finally, any potential additional computational cost observed for the RPN-BO depending on the problem setup (e.g. the output dimension) can be justified by a better optimal solution which was observed for all experiments considered in this work compared to the optimum returned by the HOGP and TuRBO methods as shown across all problems considered. Such an improvement can justify additional computational cost if it is significant and if the quality of the returned optimum is critical.

\section{Discussion}
\label{sec:cl}

We proposed a bootstrapped Randomized Prior Network approach for Bayesian Optimization (RPN-BO) that takes advantage of the neural networks ensemble's expressiveness in order to map high-dimensional spaces by not emulating the objective function directly, but rather by emulating the high-dimensional output which the objective function depends on. We also extended the proposed RPN-BO framework to the most general problem setup of constrained multi-fidelity optimization. The RPN-BO relies on reparametrized Monte Carlo (MC)-approximations for acquisition functions to obtain corresponding unbiased gradient estimators, determine the newly acquired points within the optimization process and enable parallel multi-point acquisition per optimization iteration. Such properties prove to be crucial in case of limited computational resources and/or multi-fidelity optimization tasks. 

The proposed RPN-BO approach showed higher convergence rate and returned more optimal solutions when tested against state-of-the-art High-Order Gaussian Process and TuRBO approaches across several high-dimensional Bayesian optimization benchmarks. We also tested the robustness of the RPN-BO technique with respect to the neural networks' architecture, the acquisition function, and the number of acquired points per optimization iteration.

The RPN-BO could be improved by defining a more general framework to accommodate non-differentiable and non-smooth objective functions. This limitation is inherently related to most (if not all) existing machine learning surrogate models, whose training require differentiability of the chosen metrics with respect to the model's parameters.

\section*{Author Contributions}
MAB and PP conceptualized the research. MAB, SS and MJ designed the numerical studies. MAB, MJ and RY performed the  simulations. PP and SS provided funding. MAB, MJ and PP wrote the manuscript.

\section*{Competing Interests}
The authors declare that they have no competing interests.

\section*{Acknowledgements}
We would like to acknowledge support from the US Department of Energy under the Advanced Scientific Computing Research program (grant DE-SC0019116), the US Air Force (grant AFOSR FA9550-20-1-0060), the US Department of Energy/Advanced Research Projects Agency (grant DE-AR0001201) and National Science Foundation (AGS-PRF Fellowship Award AGS-2218197). We also thank the developers of the software that enabled our research, including JAX \cite{jax2018github}, Matplotlib \cite{hunter2007matplotlib}, and NumPy \cite{harris2020array}.

\bibliographystyle{unsrt}
\bibliography{main.bib}

\clearpage
\appendix

\section{Re-parameterized Monte Carlo Approximation of Multiple-Point (Parallel) Acquisition Functions}
\label{app_sec:reparam}

The derivations of re-parameterized Monte Carlo (MC) approximations for the Likelihood-weighted Lower Confidence Bound (LW-LCB), CLSF and Output-Weighted Lower Confidence Bound with Constraints (LW-LCBC) acquisition functions are detailed in the context of Randomized Prior Networks (RPN)-based Bayesian Optimization (BO).

\subsection{Likelihood-weighted Lower Confidence Bound (LW-LCB) Acquisition Function}\label{app_subsec:LW_LCB_MC}
The following derivation can be naively repeated to derive the re-parameterized MC approximation of any other likelihood-weighted extension of an acquisition function for which a re-parameterized MC approximation is available. The output weighted Lower Confidence Bound acquisition function \cite{blanchard2021bayesian, blanchard2020informative, blanchard2020output} is defined as:
\begin{equation}\label{equ:LW_LCB}
\begin{aligned}
    a_{\textrm{LW-LCB}}(\bm{x}) = \mu(\bm{x}) - \sqrt{\kappa} w(\bm{x}) \sigma(\bm{x}), 
\end{aligned}
\end{equation}
where $\kappa$ is a user-defined confidence parameter that controls the exploration versus exploitation trade-off \cite{Wilson2018} and $\mu(\bm{x})$ and $\sigma(\bm{x})$ are the predicted mean and standard deviation respectively. The weight $w(\bm{x})$ plays a role in regularizing the uncertainty from the predicted posterior distribution and is computed via the following Gaussian mixture approximation:
\begin{equation}\label{equ:LW_LCB_weight}
\begin{aligned}
  w(\bm{x}) = \frac{p_x(\bm{x})}{p(\mu(\bm{x}))} \approx \sum_{k=1}^{n_\textit{GMM}} \alpha_k \mathcal{N}(\bm{x}; \mathbf{\gamma}_k, \Sigma_k).
\end{aligned}
\end{equation}

The single-point acquisition function (\ref{equ:LW_LCB}) can be re-written as:
\begin{equation}\label{equ:LW_LCB_weight2}
\begin{aligned}
    a_{\textrm{LW-LCB}}(\bm{x}) = \int\limits_{-\infty}^{\infty} [ \ \mu(\bm{x}) -\sqrt{\kappa\pi/2} \ w(\bm{x}) \ | \ \sigma(\bm{x}) \ z \ | \ ] \ \mathcal{N}(z;0,1) \ dz , 
\end{aligned}
\end{equation}
since the weight $w(\bm{x})$ is positive an where we used the following property:
\begin{equation}\label{equ:prop_int}
\begin{aligned}
    \sqrt{\pi/2} \int\limits_{-\infty}^{\infty} |\sigma z| \mathcal{N}(z;0,1) dz = \sigma . 
\end{aligned}
\end{equation}

With the change of variable $\gamma \equiv f(g(\bm{x})) - \mu(\bm{x}) \sim \mathcal{N}(0,\sigma(\bm{x}))$, equation (\ref{equ:LW_LCB_weight2}) can be equivalently written as follows:
\begin{equation}
\begin{aligned}
    a_{\textrm{LW-LCB}}(\bm{x}) = \int\limits_{-\infty}^{\infty} [ \ \mu(\bm{x}) -\sqrt{\kappa\pi/2} \ w(\bm{x}) \ |\gamma| \ ] \ \mathcal{N}(\gamma;0,\sigma^2(\bm{x})) \ d\gamma . 
\end{aligned}
\end{equation}

Assuming that the RPN ensemble predictions $f(\hat{g}_{\theta_i,\gamma_i}(\bm{x}))$ provide sufficiently accurate estimates for $\mu(\bm{x})$ and $\sigma(\bm{x})$ and since by definition $\gamma \equiv f(g(\bm{x})) - \mu(\bm{x})$, an MC estimate of the single-point acquisition function is obtained as follows:
\begin{equation}\label{equ:acqLWLCB_RPNs}
\begin{aligned}
    a_{\textrm{LW-LCB}}(\bm{x})\approx \frac{1}{N_s}\sum\limits_{i=1}^{N_s} \mu(\bm{x})-\sqrt{\kappa\pi/2} \ w(\bm{x}) \ |f(\hat{g}_{\theta_i,\gamma_i}(\bm{x}))-\mu(\bm{x})| \ , 
\end{aligned}
\end{equation}
where $\theta_i\sim p(\theta)$ , $\gamma_i\sim p(\gamma)$, $N_s$ is the RPN ensemble size and $\mu(\bm{x})$ is the RPNs' mean estimates of the scalar objective function evaluated at point $\bm{x}$.

Since the LW-LCB acquisition function is to be minimized, an MC estimate of the multi-point acquisition function using the RPNs' predictions can be readily obtained as follows:
\begin{equation}\label{equ:acqLWLCB_RPN}
\begin{aligned}
    a_{\textrm{LW-LCB}}(\bm{x}_1,\ldots,\bm{x}_q)\approx \\
    \frac{1}{N_s}\sum\limits_{i=1}^{N_s}\min\limits_{j=1,\ldots,q} \{ \mu(\bm{x}_j)-\sqrt{\kappa\pi/2} \ w(\bm{x}_j) \ |f(\hat{g}_{\theta_i,\gamma_i}(\bm{x}_j))-\mu(\bm{x}_j)| \} \ , 
\end{aligned}
\end{equation}
where $\theta_i\sim p(\theta)$ , $\gamma_i\sim p(\gamma)$, $N_s$ is the RPN ensemble size and $\mu(\bm{x}_j)$, $1\leq j\leq q$ are the RPNs' mean estimates of the scalar objective function evaluated at point $\bm{x}_j$.

\begin{figure}[h]
\centering
\includegraphics[width=\textwidth]{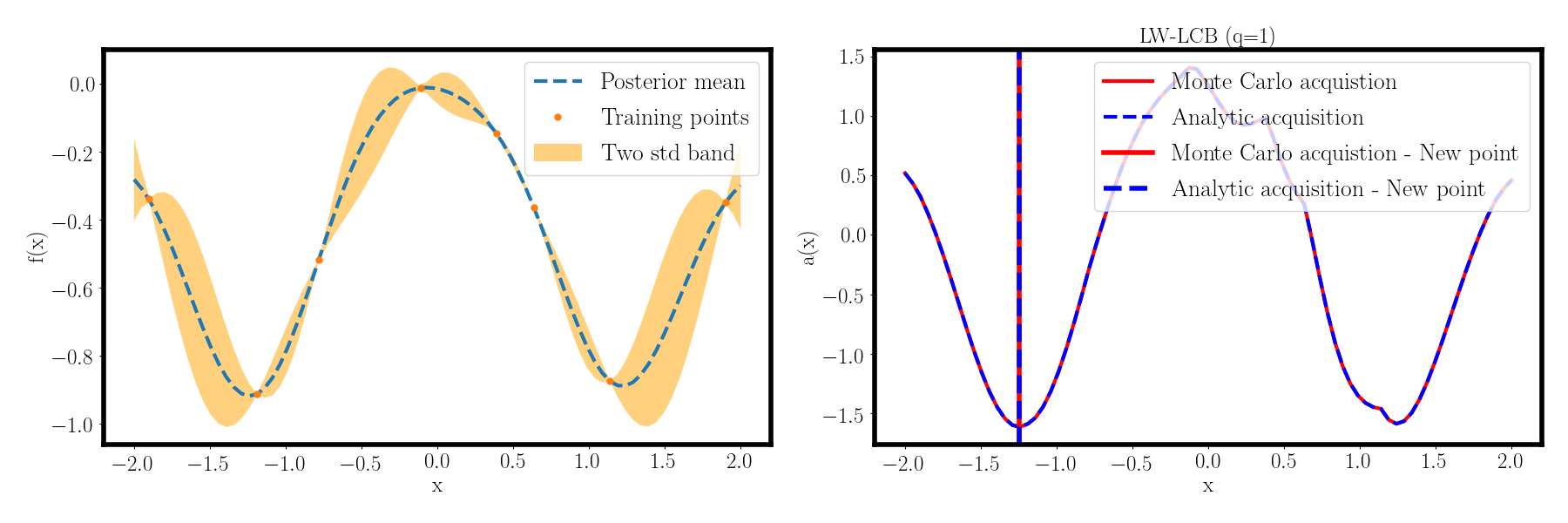}
\caption{{\em LW-qLCB criterion applied to a double well example:} (Left) Training data and posterior distribution. (right) Verification of the reparameterized MC approximation with the analytical acquisition function.}
\label{fig:LW-qLCB_q_1}
\end{figure}

\begin{figure}
\centering
\includegraphics[width=\textwidth]{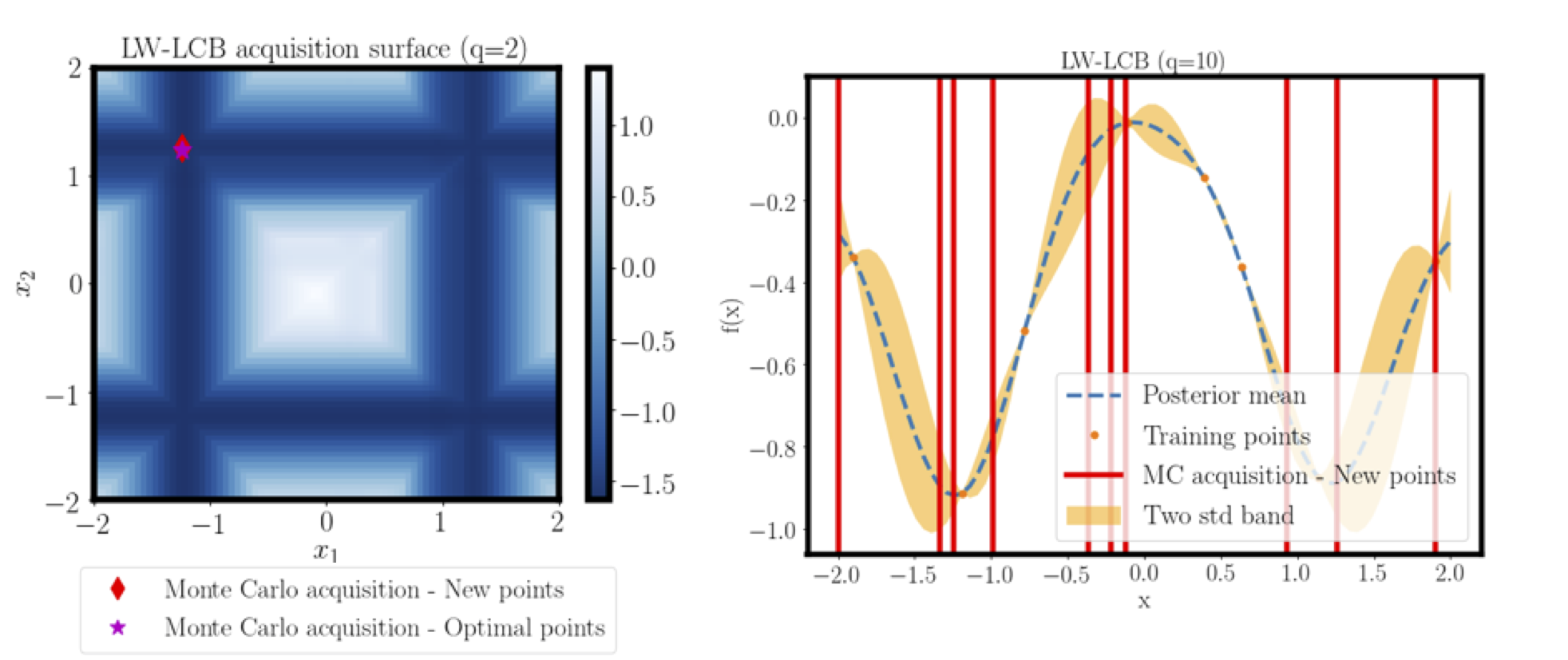}
\caption{{\em LW-qLCB criterion applied to a double well example:} (Left) Acquisition surface ($q=2$). (right) Multi-point selection with $q=10$.}
\label{fig:LW-qLCB_q_2_10}
\end{figure}

Figures \ref{fig:LW-qLCB_q_1} and \ref{fig:LW-qLCB_q_2_10} show the results obtained with the derived re-parameterized MC approximation of the LW-LCB criterion applied to a double well example. We can verify that the acquired points always contain the minima for any value of $q\in\{1,2,10\}$ that were considered.

\subsection{CLSF Acquisition Function}
The CLSF criterion is employed for learning decision boundaries. In its weighted formulation, one minimizes:
\begin{equation}\label{equ:acqCLSF_orig}
\begin{aligned}
    a_{\textrm{CLSF}}(\bm{x}) = \log \Big( \frac{|\mu(\bm{x})|^{\frac{1}{\kappa}}}{w(\bm{x})\sigma(\bm{x})+\epsilon}\Big) = \frac{1}{\kappa} \log |\mu(\bm{x})| - \log ( w(\bm{x})\sigma(\bm{x}) + \epsilon) .
\end{aligned}
\end{equation}

Alternatively, one can instead maximize:
\begin{equation}\label{equ:acqCLSF}
\begin{aligned}
    a_{\textrm{CLSF}}(\bm{x}) &= \frac{w(\bm{x})\sigma(\bm{x})}{|\mu(\bm{x})|^{\frac{1}{\kappa}} + \epsilon} .
\end{aligned}
\end{equation}

The single-point acquisition function (\ref{equ:acqCLSF}) can be re-written as:
\begin{equation}\label{equ:acqCLSF2}
\begin{aligned}
    a_{\textrm{CLSF}}(\bm{x}) = \int\limits_{-\infty}^{\infty} \frac{\sqrt{\kappa\pi/2} \ w(\bm{x})}{|\mu(\bm{x})|^{\frac{1}{\kappa}} + \epsilon} \ | \ \sigma(\bm{x}) \ z \ | \ \mathcal{N}(z;0,1) \ dz , 
\end{aligned}
\end{equation}
since the weight $w(\bm{x})$ is positive and where we used the property (\ref{equ:prop_int}).

With the change of variable $\gamma \equiv f(g(\bm{x})) - \mu(\bm{x}) \sim \mathcal{N}(0,\sigma(\bm{x}))$, equation (\ref{equ:acqCLSF2}) can be equivalently written as follows:
\begin{equation}
\begin{aligned}
    a_{\textrm{CLSF}}(\bm{x}) = \int\limits_{-\infty}^{\infty}\frac{\sqrt{\kappa\pi/2} \ w(\bm{x})}{|\mu(\bm{x})|^{\frac{1}{\kappa}} + \epsilon} \ |\gamma| \ \mathcal{N}(\gamma;0,\sigma^2(\bm{x})) \ d\gamma . 
\end{aligned}
\end{equation}

Assuming that the RPN ensemble predictions $f(\hat{g}_{\theta_i,\gamma_i}(\bm{x}))$ provide sufficiently accurate estimates for $\mu(\bm{x})$ and $\sigma(\bm{x})$ and since by definition $\gamma \equiv f(g(\bm{x})) - \mu(\bm{x})$, an MC estimate of the single-point acquisition function is obtained as follows:
\begin{equation}\label{equ:acqLWLCB_RPNs}
\begin{aligned}
    a_{\textrm{CLSF}}(\bm{x})\approx \frac{1}{N_s}\sum\limits_{i=1}^{N_s} \frac{\sqrt{\pi/2} \ w(\bm{x}_j) \ |f(\hat{g}_{\theta_i,\gamma_i}(\bm{x}_j))-\mu(\bm{x}_j)|}{ |\mu(\bm{x}_j)|^{\frac{1}{\kappa}} + \epsilon } \ , 
\end{aligned}
\end{equation}
where $\theta_i\sim p(\theta)$ , $\gamma_i\sim p(\gamma)$, $N_s$ is the RPN ensemble size and $\mu(\bm{x})$ is the RPNs' mean estimates of the scalar objective function evaluated at point $\bm{x}$.

Since the CLSF acquisition function defined here in equation (\ref{equ:acqCLSF}) is to be maximized, an MC estimate of the multi-point acquisition function using the RPNs' predictions can be readily obtained as follows:
\begin{equation}\label{equ:acqCLSF_RPN}
\begin{aligned}
    a_{\textrm{CLSF}}(\bm{x}_1,\ldots,\bm{x}_q)\approx \frac{1}{N_s}\sum\limits_{i=1}^{N_s}\max\limits_{j=1,\ldots,q} \Big\{ \frac{\sqrt{\pi/2} \ w(\bm{x}_j) \ |f(\hat{g}_{\theta_i,\gamma_i}(\bm{x}_j))-\mu(\bm{x}_j)|}{ |\mu(\bm{x}_j)|^{\frac{1}{\kappa}} + \epsilon }
    \Big\} \ , 
\end{aligned}
\end{equation}
where $\theta_i\sim p(\theta)$, $\gamma_i\sim p(\gamma)$, $N_s$ is the RPN ensemble size and  $\mu(\bm{x}_j)$, $1\leq j\leq q$ are the RPNs' mean estimates of the scalar objective function evaluated at point $\bm{x}_j$.

\subsection{Output-Weighted Lower Confidence Bound with Constraints (LW-LCBC) Acquisition Function}

The re-parameterized MC approximation of the Output-Weighted Lower Confidence Bound with Constraints (LW-LCBC) acquisition function is presented here. Note that the derivation for the vanilla Lower Confidence Bound with Constraints acquisition function (LCBC) can be directly recovered by setting the weights equal to one. The LW-LCBC acquisition function can be formulated as follows:
\begin{equation}\label{equ:LW_LCBC}
\begin{aligned}
    a_{\textrm{LW-LCBC}}(\bm{x}) = (a_{\textrm{LW-LCB}}(\bm{x}) - \delta)\prod_{k=1}^{K}\textrm{Pr}(\mathcal{C}_{k}(\bm{x})\geq 0| \mathcal{D}) ,
\end{aligned}
\end{equation}
\noindent where $a_{\textrm{LW-LCB}}(\cdot)$ is detailed in Eq (\ref{equ:LW_LCB}), $\delta=3$ is a threshold parameter to make sure that the first term is roughly less than $0$ and $\textrm{Pr}(\mathcal{C}_{k}| \mathcal{D}) $ is the inferred posterior distribution for the constraint $\mathcal{C}_{k}$, $1\leq k\leq K$ given the observed data $\mathcal{D}$. The re-parameterization of the first term $a_{\textrm{LW-LCB}}(\cdot) - \delta$ is straightforward given the re-parameterization of $a_{\textrm{LW-LCB}}(\cdot)$ detailed in section \ref{app_subsec:LW_LCB_MC}. The re-parameterization of the constraints term can be carried out using an indicator function $\mathds{1}$, such that the final RPN-based re-parameterized MC approximation is given by:
\begin{equation}\label{equ:LW_LCBC_X}
\begin{aligned}
    a_{\textrm{LW-LCBC}}(\bm{X}) \approx \\
    \Big[\frac{1}{N_s}\sum\limits_{i=1}^{N_s}\min\limits_{j=1,\ldots,q} \{ \mu(\bm{x}_j)-\delta-\sqrt{\kappa\pi/2} \ w(\bm{x}_j) \ |f(\hat{g}_{\theta_i,\gamma_i}(\bm{x}_j))-\mu(\bm{x}_j)| \}\Big] \times \\
    \prod_{k=1}^K \frac{1}{N_s}\sum_{i=1}^{N_s} \max\limits_{j=1,\ldots,q} \mathds{1}\{c_k(\hat{h}_{k,\theta^c_{k,i},\gamma^c_{k,i}}(\bm{x}_j))\geq 0\} \ , 
\end{aligned}
\end{equation}
\noindent where $\theta_i\sim p(\theta) \ , \gamma_i\sim p(\gamma) \ , \ , \theta^c_{k,i}\sim p(\theta^c_k) \ , \gamma^c_{k,i}\sim p(\gamma^c_k)$, $N_s$ is the RPN ensemble size and  $\mu(\bm{x}_j)$, $1\leq j\leq q$ are the RPNs' mean estimates of the scalar objective function evaluated at point $\bm{x}_j$. As defined in Eq (\ref{equ:CBO}), $\hat{h}_{k,\theta^c_{k,i},\gamma^c_{k,i}}$, $1\leq k\leq K$, $1\leq i\leq N_s$, refer to the RPN surrogate models that are built to approximate the constraint black-box functions $h_{k}$, $1\leq k\leq K$, with vectorial output, and $c_k$ , $1\leq k\leq K$, refer to the scalar-output constraint functions that are assumed to be known (similar to the function $f(\cdot)$ that is defined for the objective). Note that in practice, the sigmoid function is considered as a smooth indicator function $\mathds{1}$.

\section{Hyperparameters for Numerical Experiments}
\label{app_sec:res}

For all numerical examples, the fraction $e$ of data used to train each neural network is fixed to $0.8$ as justified in \cite{Yang2022Scalable} and neural networks are trained using a learning rate equal to $10^{-3}$ with a decay rate of $0.999$ after each $10^3$ iterations. Acquisition functions optimization is always performed with 500 restarts in all numerical examples and the best optimal solution among the 500 ones that are found is taken as final solution based on the predicted acquisition function evaluation with the surrogate RPN.

\subsection{Environmental Model Function}

For the Environmental Model Function example, both MLP- and DON-based RPN ensemble size is taken as $128$ \cite{Yang2022Scalable}. The MLP-based RPNs are built by considering fully connected neural networks with $4$ hidden layers of $64$ neurons each. For the DON-based RPNs, each of the branch and trunk networks is a fully connected network with $2$ hidden layers of $64$ neurons each. Both MLP and DON ensemble neural nets are trained for $5\times10^3$ iterations. This hyper-parameter setting is chosen in order to avoid overfitting as we are interested in BO problems with high-dimensional outputs and sparse training dataset. 

\subsection{Brusselator PDE Control Problem}
\label{sec:PDE_control}

For the Brusselator PDE Control problem, both MLP- and DON-based RPN ensemble size is taken as $128$ \cite{Yang2022Scalable}. The MLP-based RPNs are built by considering fully connected neural networks with $2$ hidden layers of $64$ neurons each. For the DON-based RPNs, each of the branch and trunk networks is a fully connected network with $2$ hidden layers of $64$ neurons each. Both MLP and DON ensemble neural nets are trained for $10^4$ iterations. This hyper-parameter setting is chosen in order to avoid overfitting as we are interested in BO problems with high-dimensional outputs and sparse training dataset 

\subsection{Optical Interferometer Alignment}

To convert the problem of optimizing visibility into a BO task rather than a reinforcement learning one as defined in the original work \cite{Sorokin2020}, the simulator implemented in \url{https://github.com/dmitrySorokin/interferobotProject} is reset to a default value $(10^{-4},10^{-4},10^{-4},10^{-4})$ each time the problem is queried, then the alignment of the two mirrors is optimized starting from that setting as carried out in \cite{maddox2021bayesian} in order to make a fair comparison between the RPN-BO and HOGP methods. 

MLP- and DON-based RPN ensemble sizes are taken as $32$ and $16$, respectively  \cite{Yang2022Scalable}. The MLP-based RPNs are built by considering fully connected neural networks with $2$ hidden layers of $64$ neurons each. For the DON-based RPNs, each of the branch and trunk networks is a fully connected network with $2$ hidden layers of $32$ neurons each. Both MLP and DON ensemble neural nets are trained for $5\times10^3$ iterations. This hyper-parameter setting is chosen in order to avoid overfitting as we are interested in BO problems with high-dimensional outputs and sparse training dataset. 

\subsection{Constrained Multi-fidelity Optimization of Compressor’s Blades Shape Design}


For this example, MLP-based RPN ensemble size is equal to $128$ \cite{Yang2022Scalable}. The low-fidelity MLP-based RPNs are built by considering fully connected neural networks with $2$ hidden layers of $128$ neurons each. The high-fidelity MLP-based RPNs are built by considering fully connected neural networks with $2$ hidden layers of $64$ neurons each. Finally, MLP ensemble neural nets are trained for $5\times10^3$ iterations. This hyper-parameter setting is chosen in order to avoid overfitting as we are interested in BO problems with high-dimensional outputs and sparse training dataset. 

\end{document}